\documentclass[letterpaper, 10 pt, conference]{ieeeconf}  

\IEEEoverridecommandlockouts                              

\title{\LARGE \bf
	Exploring Domain Shift on Radar-Based 3D Object Detection Amidst Diverse Environmental Conditions
}

\author{Miao Zhang$^{1,2}$, Sherif Abdulatif$^{1}$, Benedikt Loesch$^{1}$, Marco Altmann$^{1}$, Marius Schwarz$^{1}$ and Bin Yang$^{2}$
	\thanks{$^{1}$Robert Bosch GmbH, Corporate Research, 71272, Germany.
		{\tt \{miao.zhang5, sherif.abdulatif, benedikt.loesch, marco.
			altmann, marius.schwarz\}@de.bosch.com}}%
	\thanks{$^{2}$University of Stuttgart, Institute of Signal Processing and System Theory, 70569, Germany.
		{\tt bin.yang@iss.uni-stuttgart.de}}%
}

\renewcommand{\baselinestretch}{0.965}
\usepackage{url}
\usepackage{xcolor}
\usepackage{booktabs}
\usepackage{graphicx}
\usepackage{tikz}
\usepackage{xcolor}
\usepackage{subcaption}
\usepackage{multirow}
\usepackage{threeparttable}
\usepackage[T1]{fontenc}
\usepackage[utf8]{inputenc}
\usepackage{pgfplots}
\usepackage{grffile}
\usepackage{array}
\let\labelindent\relax
\usepackage{enumitem}
\pgfplotsset{compat=newest}
\usetikzlibrary{plotmarks}
\usetikzlibrary{arrows.meta}
\usepgfplotslibrary{patchplots}
\definecolor{yellowbox}{RGB}{242, 133, 0}
\definecolor{greenbox}{RGB}{0, 200, 50}

\makeatletter
\let\NAT@parse\undefined
\makeatother
\usepackage[colorlinks=true, linkcolor=black, citecolor=black, filecolor=black, urlcolor=black]{hyperref}

\newcommand{\redline}{\textcolor{red}{\rule[0.5ex]{1.5em}{1.5pt}}}
\newcommand{\greenline}{\textcolor{greenbox}{\rule[0.5ex]{1.5em}{1.5pt}}}
\newcommand{\yellowline}{\textcolor{yellowbox}{\rule[0.5ex]{1.5em}{1.5pt}}}
\begin{document}
	
	\maketitle
	\thispagestyle{empty}
	\pagestyle{empty}

	\begin{abstract}
		The rapid evolution of deep learning and its integration with autonomous driving systems have led to substantial advancements in 3D perception using multimodal sensors. Notably, radar sensors show greater robustness compared to cameras and lidar under adverse weather and varying illumination conditions. This study delves into the often-overlooked yet crucial issue of domain shift in 4D radar-based object detection, examining how varying environmental conditions, such as different weather patterns and road types, impact 3D object detection performance. Our findings highlight distinct domain shifts across various weather scenarios, revealing unique dataset sensitivities that underscore the critical role of radar point cloud generation. Additionally, we demonstrate that transitioning between different road types, especially from highways to urban settings, introduces notable domain shifts, emphasizing the necessity for diverse data collection across varied road environments. To the best of our knowledge, this is the first comprehensive analysis of domain shift effects on 4D radar-based object detection. We believe this empirical study contributes to understanding the complex nature of domain shifts in radar data and suggests paths forward for data collection strategy in the face of environmental variability.
	\end{abstract}

	\section{INTRODUCTION}
	With the rapid development of deep learning technology, various sensors are utilized to optimize autonomous driving systems. Data inputs such as camera images and lidar-generated point clouds are commonly integral to modern advanced 3D perception systems. While using a deep learning-based model with the input of a camera and lidar point cloud can achieve significant perception results~\cite{liang2022bevfusion, shi2022pillarnet}, these systems still face challenges related to illumination, adverse weather conditions, and high cost~\cite{srivastav2023radars}. In contrast, radar sensors offer a more robust signal at an acceptable cost under such conditions. Additionally, the latest 4D radar not only provides high-resolution range and velocity information but also includes azimuth and elevation angle data. Considering its advantages, radar sensors are increasingly recognized as a promising avenue in the 3D perception field.
	
\begin{figure}[tbp]
		\vspace{0.5em}
		\centering
		\includegraphics[width=\columnwidth]{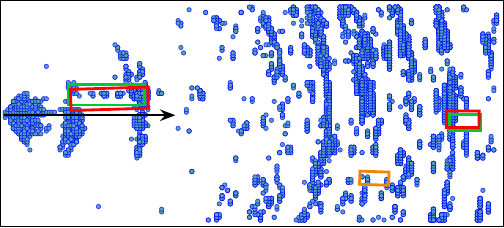}
		\caption{\textbf{Bird's eye view of radar object detection in heavy snow conditions from K-Radar dataset~\cite{paek2022k}:} The \greenline{} boxes are ground truth, \redline{} boxes are predictions from the model trained on all-weather data, and \yellowline{} boxes are predictions from the model trained only on normal weather conditions.}
		\label{fig: pcs}
		\vspace{-1.4em}
	\end{figure}

	Although radar signals are relatively robust under adverse weather conditions compared to optical-based signals (lidar and camera), their features can still vary with different environmental conditions~\cite{appiah2023object,mohammed2020perception}. Research has shown that most radar datasets are collected under normal weather conditions~\cite{paek2022k}. A model trained solely on these datasets may yield unreliable predictions under different weather conditions, as shown in Fig.~\ref{fig: pcs}. This phenomenon is known as ``domain shift'' in the machine learning field: it occurs when a model trained on one set of data (the source domain) encounters different data distributions (the target domain), potentially reducing its performance. For an open-world task such as autonomous driving, it is crucial to know if a considerable domain shift happens and utilize domain adaptation methods to solve that accordingly. Recent studies have begun to address the impact of domain shift on 3D object detection, primarily focusing on lidar sensors~\cite{vattem2022rethinking,eskandar2024empirical}. However, the specific effects of domain shift on radar object detection remain less explored, especially given the unique characteristics of radar sensors.
	
	In this paper, we conduct a comprehensive empirical study into the effects of domain shift on 4D radar-based object detection across various environmental conditions, including different weather scenarios and road types. Our research demonstrates that domain shifts occur consistently under various weather conditions, although their impacts vary significantly across different datasets. This emphasizes the crucial role of generating radar point clouds. Additionally, we found that neither different network architectures nor increasing the size of datasets substantially mitigates the impact of weather-related domain shifts, highlighting the importance of domain adaptation techniques. We also observed domain shift effects when transitioning between different road types, particularly from highways to urban settings. This underscores the importance of gathering a diverse range of data from various road environments to enhance the effectiveness of radar-based 3D object detection.
	\section{RELATED WORK}
	\subsection{Weather Effect on Radar Signals}
	The radar sensor transmits electromagnetic waves and captures the reflected signal to determine the range, angle, and velocity information of the target. Compared to lidar, radar is more robust in adverse weather conditions, given the relatively long wavelength. Nevertheless, there are still some challenges like interference, attenuation, and backscattering, which can degrade the quality of the signal: Courova \emph{et~al.}~\cite{gourova2017analysis} studied the impact of rain on the performance of automotive radar, showing that the attenuation and clutter during rain can reduce target detectability. The weather effect on radar signals was also mentioned in a recent work~\cite{appiah2023object}: the precipitation condition (e.g., rain and snow) can have a significant impact on the signal, whereas conditions with airborne particles (e.g., dusty storm and smoke) have minimal effect on the radar signals. Most of this research focuses on the characteristics of radar signals, with less emphasis on the performance gap in deep learning-based perception. 
	From a deep learning perspective, RADIATE~\cite{sheeny2021radiate} addresses domain shifts in adverse weather conditions, but it is constrained by an inconsistent and limited dataset size and only offers 2D evaluation. Unlike previous works, our study aims to conduct a comprehensive analysis of domain shifts for 4D radar-based object detection across various weather conditions.
	
	\subsection{4D Radar Datasets}\label{sec: 4D data}
	The most recent 4D radar sensor provides additional elevation angle information, enriching the traditional 3D data (range, velocity, and azimuth angle) previously available from its predecessors. Given the improved angular resolution and range capabilities, 4D radar sensors are garnering increasing interest in autonomous driving. However, the availability of comprehensive open-source 4D radar datasets remains limited. Common datasets in 3D object detection, like Waymo~\cite{sun2020scalability} and KITTI~\cite{geiger2012kitti}, only provide lidar point clouds. Additionally, datasets like NuScenes~\cite{caesar2020nuscenes}, CRUW~\cite{wang2021rethinking}, and RADIATE~\cite{sheeny2021radiate} offer radar point clouds captured by 3D radar sensors.
	Astyx was the first to provide high-resolution 4D radar point clouds, but its dataset size is quite limited~\cite{meyer2019automotive}. Since then, several 4D radar datasets have been released, but they are not very suitable for domain shift investigations due to a lack of abundant data from various weather conditions and diverse scenes, such as VOD~\cite{palffy2022multi}, TJ4RAD~\cite{Zheng_2022}, and aiMotive~\cite{matuszka2022aimotive}.
	Recently, a 4D radar dataset named K-Radar, which includes data from different adverse weather conditions, was published~\cite{paek2022k}. It contains over 35k frames from 58 recordings, capturing data across seven types of weather conditions.
	
	\subsection{Domain Shift on 3D Object Detection}
	Domain shift has been extensively researched in various tasks, such as semantic segmentation~\cite{sankaranarayanan2018learning}, image classification~\cite{zhang2021adaptive}, and 2D object detection~\cite{2ddomain2022}. Beyond camera sensors, pioneering works have also explored 3D object detection with lidar sensors. For instance, Vattem \emph{et~al.}~\cite{vattem2022rethinking} examined the domain shift effect in adverse weather conditions, demonstrating that a model trained in clear weather can achieve competitive results in adverse conditions. Wang \emph{et~al.}~\cite{wang2020train} highlighted geographic domain shift by training a model on data from Germany and testing it on data from the USA. A recent empirical study~\cite{eskandar2024empirical} provided a comprehensive analysis of lidar 3D object detection, considering factors such as model architectures, location domains, and weather domains, showing that training on clear weather samples yields more robust results than training on adverse weather samples. However, the impact of domain shift varies significantly across different sensors, suggesting that findings specific to lidar may not be directly applicable to radar sensors. To our knowledge, we are the first to study the domain shift impact on radar 3D object detection. In this paper, our research will focus exclusively on environmental conditions like weather and road type. Other aspects, such as model structure or sensor type domain gaps, are beyond the scope of this study.
	\section{METHODOLOGY}
	\begin{table}[bp]
		\vspace{-1em}
		\caption{
			Summary of datasets, covering radar specifications, environmental diversity, and training/testing split.} 
		\label{tab:dataset}
		\resizebox{\columnwidth}{!}{
			\begin{threeparttable}
				\centering
				\begin{tabular}{lcr} 
					\toprule 
					Dataset & K-Radar & Bosch-Radar\\
					\midrule 
					Radar& $1\times4D$& $5\times4D$\\
					ROI$^{1}$ (x,y,z) &\scriptsize{[(0, 72), (-16, 16), (-2, 7.6)]} & \scriptsize{ [(0, 80), (-20, 20), (-10, 10)]}\\
					Weather types&7&3\\
					Cities&1&8\\
					Frames$^{2}$ (train/test)&27,024/7,970&180,000/65,554\\
					Seqs$^{2}$ (train/test)&44/14&1,146/312 \\
					\bottomrule 
				\end{tabular}
				\begin{tablenotes}
					\footnotesize
					\item[1] Region of Interest
					\item[2] Here we give the size of all the data. During the experiment, we will use different subsets for different environmental conditions. 
				\end{tablenotes}
			\end{threeparttable}
		}
	\end{table}
	
	\subsection{Dataset}
	\label{sec: dataset}
	As mentioned in Sec.~\ref{sec: 4D data}, most public 4D radar datasets are unsuitable for investigating environmental domain shifts. However, in this study, we will use one suitable public dataset along with a self-collected dataset to conduct a comprehensive analysis. Detailed information about the datasets is provided below, with specific data outlined in Table~\ref{tab:dataset}:
	\begin{itemize}[wide = 0.1em]
		\item \textbf{K-Radar dataset~\cite{paek2022k}:} This dataset was collected in South Korea using a single 4D radar sensor. It includes point clouds from seven different weather conditions: Normal, Rain, Heavy Snow, Light Snow, Overcast, Sleet, and Fog, with over 35k frames in total. The original K-Radar dataset has a default splitting configuration where the training and test sets contain frames from the same recordings, potentially leading to biased validation outcomes due to the short duration of the recordings and similarity of the frames. To address this, we re-split the dataset to ensure no recording overlap between the training and test sets. This approach prevents any bias and ensures a more reliable evaluation of domain shift. Additionally, we maintained the same dataset size for different settings to allow for fair comparison. The training set now consists of 27,024 frames, while the test set includes 7,970 frames, covering all adverse weather conditions.
		\item \textbf{Bosch-Radar dataset:} In addition to evaluating a public dataset, we conducted experiments on a private dataset, referred to as the Bosch-Radar dataset. As shown in Table \ref{tab:dataset}, this dataset is significantly larger than the K-Radar dataset, containing over 180k frames collected from eight different cities in Germany over 27 days (19 days in the training set and 8 in the test set). Using five 4D radar sensors at the front view, we acquired high-quality radar point clouds. Our dataset includes three weather conditions (Normal, Overcast, and Rain) and three types of road environments (Urban, Rural, and Highway). To examine the domain shift effect across different dataset scales, we organized training sets of varying sizes (60k, 20k, and 10k frames). Given the broad scope and variety within our dataset, we anticipate obtaining robust and convincing results.
	\end{itemize}
	\begin{figure}[b]
		\vspace{-1em}
		\centering
		\includegraphics[width=1\linewidth]{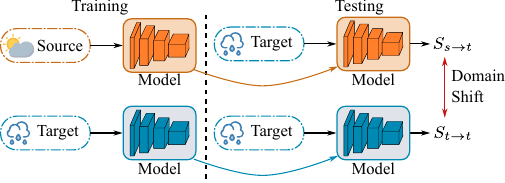}
		\caption{Illustration of the experimental workflow highlighting domain shift analysis between source and target datasets.}
		\label{fig:block}
	\end{figure}
	\vspace{-1mm}\subsection{Implementation Details}
	\label{sec: implement}
	To conduct a comprehensive study, we validate the domain shift effect using two representative 3D object detectors, RTNH~\cite{paek2022k} and SECOND~\cite{SECOND}, on the aforementioned datasets. Our approach involves training the models with data from different domains and validating them on a fixed test set. To ensure a fair comparison for weather condition experiments, we maintain an identical number of training frames. Specifically, we use the bottleneck value of 8,764 frames from the normal samples as the training set size for each scenario in the K-Radar dataset (normal weather and all-weather data). For our Bosch-Radar dataset, we organize the data into three different scales, as previously mentioned. For road-type experiments, we maintain a consistent number of objects (200k) to account for variations in the average number of objects across different road types. For instance, there is an average of 8.12 sedans per frame in urban recordings, compared to only 4.10 per frame on highways in the Bosch-Radar dataset.
	
	The models are implemented using the OpenPCDet framework~\cite{openpcdet2020} and PyTorch. In line with K-Radar~\cite{paek2022k}, we utilize the AdamW optimizer and Cosine Annealing learning rate scheduler. Detection performance is measured using the $AP_{3D}$ and $AP_{BEV}$ metrics from the KITTI dataset~\cite{geiger2012kitti}. We run each experiment three times on a NVIDIA A100 GPU, and the mean value along with the standard deviation is reported in the form of $\textbf{mean} \pm \scriptstyle{std}$.
	
	By observing the performance gap across different domains, we can determine whether a domain shift exists, as illustrated in Fig.~\ref{fig:block}. For example, if a model trained on the source domain $s$ performs similarly in the target domain $t$ as a model trained on the target domain, we can conclude that there is no significant domain shift from $s$ to $t$, or that the two domains substantially overlap. It is important to focus primarily on comparing the performance gaps in the target domain between models trained on different domains, rather than on the absolute performance of each model.
	
	Assuming a model trained on domain $s$ yields a validation score of $S_{s \rightarrow t}$ (e.g., Average Precision) when tested on domain $t$ and a score of $S_{s \rightarrow s}$ when tested on domain $s$, the relationship between $S_{s \rightarrow s}$ and $S_{s \rightarrow t}$ alone does not necessarily indicate a domain shift from $s$ to $t$. To confirm a domain shift, we need to compare the original performance on the target domain $S_{t \rightarrow t}$ with the shift performance $S_{s \rightarrow t}$. If $S_{t \rightarrow t} \gg S_{s \rightarrow t}$, then a domain shift occurs from $s$ to $t$. Conversely, if $S_{t \rightarrow t} \approx S_{s \rightarrow t}$, then the two domains overlap significantly, or $t$ is a subdomain of $s$.
	
	\section{Weather Domain Shift}
	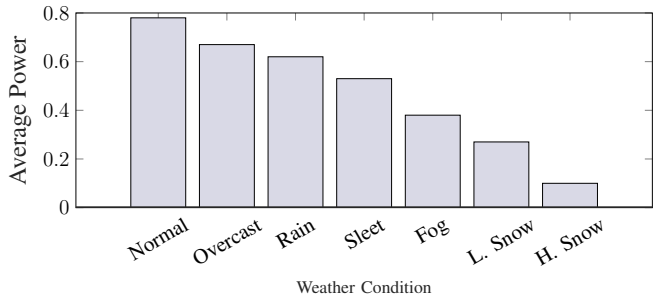
\begin{figure}[tbp]
		\vspace{0.7em}
		\centering
		{\resizebox{\columnwidth}{!}{
%
%
\definecolor{mycolor1}{rgb}{0.85000,0.85000,0.90000}%
\begin{tikzpicture}

\begin{axis}[%
width=4.521in,
height=1.528in,
at={(0.758in,1.219in)},
scale only axis,
bar shift auto,
xmin=-0.2,
xmax=8.2,
xtick={1,2,3,4,5,6,7},
xticklabels={{Normal},{Overcast},{Rain},{Sleet},{Fog},{L. Snow},{H. Snow}},
xticklabel style={rotate=30},
xlabel style={font=\color{white!15!black}},
xlabel={Weather Condition},
ymin=0,
ymax=0.8,
yticklabel style = {font=\large},
xticklabel style = {font=\large},
ylabel style={font=\color{white!15!black}},
ylabel={\Large Average Power},
axis background/.style={fill=white}
]
\addplot[ybar, bar width=0.8, fill=mycolor1, draw=black, area legend] table[row sep=crcr] {%
1	0.78\\
2	0.67\\
3	0.62\\
4	0.53\\
5	0.38\\
6	0.27\\
7	0.1\\
};
\addplot[forget plot, color=white!15!black, line width=1.5pt] table[row sep=crcr] {%
-0.2	0\\
8.2	0\\
};
\end{axis}
\end{tikzpicture}%
}}
		\caption{Average power of received radar signal for different weather conditions. Notations: \textbf{L.~Snow} represents Light Snow, and \textbf{H.~Snow} denotes Heavy Snow.}
		\label{fig: weather_info}
		\vspace{-5mm}
	\end{figure}
	
	\subsection{Results on K-Radar}
	On the K-Radar dataset, we investigate the effect of weather domain shifts by comparing the results under two different settings: one model is trained exclusively on normal weather data, while the other is trained on a mixed dataset encompassing all seven weather conditions, with both datasets being of identical size. We follow a two-class object detection setting from the wide field of view (FOV) configuration of the original K-Radar, where one class is ``sedan'' and the other is ``bus or truck''. The evaluation results for two networks are concluded in Table~\ref{tab: kradar}. Due to the non-overlapping split setting mentioned in Sec.~\ref{sec: dataset}, the performance may be worse than reported in the original paper~\cite{paek2022k}. Our findings are detailed as follows:
	
	\begin{table*}[ht]
		\vspace{0.5em}
		\centering
		\caption{
			Comparison of $AP_{BEV}$/$AP_{3D}$ metrics for weather domain shift (source$\rightarrow$target) on the new K-Radar split.}
		\label{tab: kradar}
		\resizebox{\textwidth}{!}{
			\begin{threeparttable}
				\setlength{\extrarowheight}{0.8pt}
				\begin{tabular}{l|cc|cc}
					\toprule
					Networks&\multicolumn{2}{c|}{RTNH\cite{paek2022k}}&\multicolumn{2}{c}{SECOND\cite{SECOND}}\\
					\hline
					Sedan & Normal$\rightarrow$ & Mixed$\rightarrow$&Normal$\rightarrow$ & Mixed$\rightarrow$ \\
					\cline{1-5}
					$\rightarrow$Normal & $\textbf{35.44} \pm \scriptstyle{0.33}/\textbf{31.94}\pm \scriptstyle{0.40}$ & $\textbf{32.77} \pm \scriptstyle{1.44}/\textbf{32.27}\pm \scriptstyle{0.48}$ &$\textbf{34.82} \pm \scriptstyle{1.14}/\textbf{26.96}\pm \scriptstyle{1.38}$&$\textbf{34.45} \pm \scriptstyle{0.69}/\textbf{27.09}\pm \scriptstyle{0.35}$\\
					$\rightarrow$Rain & \textcolor{red}{$\textbf{35.54} \pm \scriptstyle{0.19}/\textbf{23.37}\pm \scriptstyle{0.33}$}& $\textbf{48.12} \pm \scriptstyle{0.89}/\textbf{40.61}\pm \scriptstyle{3.34}$ &\textcolor{red}{$\textbf{37.41} \pm \scriptstyle{0.60}/\textbf{23.76}\pm \scriptstyle{2.31}$}&$\textbf{48.29} \pm \scriptstyle{0.66}/\textbf{37.17}\pm \scriptstyle{0.43}$\\
					$\rightarrow$Heavy snow & \textcolor{red}{ $\textbf{11.03} \pm \scriptstyle{2.19}/\textbf{0.00}\pm \scriptstyle{0}$}& $\textbf{40.12} \pm \scriptstyle{3.62}/\textbf{17.75}\pm \scriptstyle{0.61}$&\textcolor{red}{$\textbf{6.58} \pm \scriptstyle{1.19}/\textbf{0.76}\pm \scriptstyle{1.07}$}&$\textbf{39.09} \pm \scriptstyle{0.81}/\textbf{27.05}\pm \scriptstyle{0.52}$ \\
					$\rightarrow$Light snow &  \textcolor{red}{$\textbf{18.26} \pm \scriptstyle{0.39}/\textbf{17.84}\pm \scriptstyle{0.28}$}& $\textbf{82.99} \pm \scriptstyle{5.23}/\textbf{81.72}\pm \scriptstyle{4.01}$&\textcolor{red}{$\textbf{20.74} \pm \scriptstyle{0.54}/\textbf{20.59}\pm \scriptstyle{0.32}$}&$\textbf{87.91}\pm \scriptstyle{0.44}/\textbf{87.24} \pm \scriptstyle{1.39}$ \\
					$\rightarrow$Overcast & $\textbf{37.06} \pm \scriptstyle{2.31}/\textbf{30.18}\pm \scriptstyle{0.89}$& $\textbf{34.72} \pm \scriptstyle{1.08}/\textbf{32.28}\pm \scriptstyle{1.05}$&$\textbf{34.72} \pm \scriptstyle{0.30}/\textbf{25.22}\pm \scriptstyle{4.07}$&$\textbf{35.58} \pm \scriptstyle{1.60}/\textbf{29.66}\pm \scriptstyle{4.91}$ \\
					$\rightarrow$Sleet & \textcolor{red}{$\textbf{55.13} \pm \scriptstyle{1.45}/\textbf{21.00}\pm \scriptstyle{0.80}$}&$\textbf{75.34} \pm \scriptstyle{0.40}/\textbf{53.91}\pm \scriptstyle{4.16}$ &\textcolor{red}{$\textbf{57.46} \pm \scriptstyle{1.49}/\textbf{23.74}\pm \scriptstyle{4.00}$}&$\textbf{74.16} \pm \scriptstyle{0.72}/\textbf{58.15}\pm \scriptstyle{2.70}$ \\
					$\rightarrow$Fog &$\textcolor{red}{\textbf{71.95} \pm \scriptstyle{1.64}/\textbf{53.18}\pm \scriptstyle{6.50}}$&$\textbf{79.70} \pm \scriptstyle{0.69}/\textbf{67.09}\pm \scriptstyle{2.67}$&\textcolor{red}{$\textbf{61.00} \pm \scriptstyle{5.93}/\textbf{42.73}\pm \scriptstyle{6.07}$}&$\textbf{77.81} \pm \scriptstyle{2.03}/\textbf{60.27}\pm \scriptstyle{3.88}$  \\
					\cline{1-5}
					Bus or Truck & Normal$\rightarrow$ & Mixed$\rightarrow$&Normal$\rightarrow$& Mixed$\rightarrow$\\
					\cline{1-5}
					$\rightarrow$Normal &  $\textbf{23.46} \pm \scriptstyle{0.77}/\textbf{24.18}\pm \scriptstyle{0.29}$& $\textbf{25.20} \pm \scriptstyle{1.23}/\textbf{24.88}\pm \scriptstyle{1.19}$&$\textbf{29.53} \pm \scriptstyle{0.05}/\textbf{26.26}\pm \scriptstyle{0.84}$&$\textbf{27.09} \pm \scriptstyle{0.25}/\textbf{26.07}\pm \scriptstyle{0.30}$ \\
					$\rightarrow$Heavy snow & \textcolor{red}{\textbf{0.00/0.00}} &  $\textbf{39.26} \pm \scriptstyle{4.17}/\textbf{2.26}\pm \scriptstyle{1.38}$ &\textcolor{red}{\textbf{0.00/0.00}}&$\textbf{49.39} \pm \scriptstyle{6.28}/\textbf{1.46}\pm \scriptstyle{1.15}$ \\
					$\rightarrow$Light snow & \textcolor{red}{$\textbf{26.03} \pm \scriptstyle{4.35}/\textbf{12.92}\pm \scriptstyle{4.54}$} & $\textbf{67.59} \pm \scriptstyle{0.46}/\textbf{35.94}\pm \scriptstyle{5.98}$&\textcolor{red}{$\textbf{26.03} \pm \scriptstyle{4.40}/\textbf{6.23}\pm \scriptstyle{1.37}$}&$\textbf{75.79} \pm \scriptstyle{7.21}/\textbf{53.38}\pm \scriptstyle{3.16}$  \\
					$\rightarrow$Overcast & $\textbf{17.82}\pm \scriptstyle{0.36}/\textbf{17.66} \pm \scriptstyle{0.52}$ & $\textbf{15.29} \pm \scriptstyle{0.55}/\textbf{15.21}\pm \scriptstyle{0.62}$ &$\textbf{26.90}\pm \scriptstyle{0.24}/\textbf{26.60} \pm \scriptstyle{0.38}$&\textcolor{red}{$\textbf{16.34} \pm \scriptstyle{0.71}/\textbf{16.07}\pm \scriptstyle{0.33}$}\\
					$\rightarrow$Sleet &\textcolor{red}{\textbf{0.00/0.00}}& $\textbf{16.47} \pm \scriptstyle{0.61}/\textbf{10.73}\pm \scriptstyle{1.67}$&\textcolor{red}{\textbf{0.00/0.00}}& $\textbf{20.17} \pm \scriptstyle{1.11}/\textbf{18.40}\pm \scriptstyle{0.53}$\\
					\bottomrule
				\end{tabular}
				\begin{tablenotes}
					\item {Note: Noticeable performance drop ($5\%+$) is highlighted in \textcolor{red}{RED}  (\textbf{Comparison across columns}).}
				\end{tablenotes}
		\end{threeparttable}}
		\vspace{-5mm}
	\end{table*}
	
	\begin{enumerate}[label=\arabic*., wide=0.1em, itemsep=0.25em]
		\item \textit{A significant domain gap is evident across five weather conditions: Rain, Heavy Snow, Light Snow, Sleet, and Fog.} 
		The snow scenario, in particular, exhibits the most severe performance degradation. Specifically, performance becomes invalid for the sedan class under heavy snow conditions according to the $AP_{3D}$ metric, and the performance drop in light snow exceeds 60\%. An examination of the average power of the point cloud reveals that the average power in snow conditions is significantly lower than in normal weather conditions, as shown in Fig.~\ref{fig: weather_info}. This reduction in power may be attributed to sensor blockages caused by ice or snow. For the bus or truck class, the two models fail under heavy snow and sleet conditions when trained only in normal weather. One possible reason is the difference in average power. Another reason could be that the bus or truck class is more challenging to detect than the sedan class due to the larger size of the objects and fewer samples. For other precipitation conditions, such as rain and sleet, the domain shift effect is still very noticeable for both models. This may be due to the cluttering and backscattering effects, which should be effectively addressed.
		
		\item \textit{Performance under overcast conditions is not severely affected by these weather domain variations.} Notably, the two normal-trained models outperform the mixed-trained model on the overcast test data, especially the SECOND model for bus or truck class, indicating a considerable domain overlap between normal and overcast weather conditions. This phenomenon also aligns with our understanding, where the illumination conditions have a minimal impact on millimeter wave propagation. 
		
		\item \textit{Comparing the effects on the two networks, we observe that although performance may vary, both experience the same domain shift effect.} When a domain shift occurs in one network, it also occurs in the other. This demonstrates that the domain shift effect related to weather conditions is quite model-agnostic and should be addressed accordingly.
	\end{enumerate}

	\subsection{Results on Bosch-Radar Dataset}
\begin{figure*}[t!]
		\vspace{0.5em}
		\captionsetup[subfigure]{justification=centering}
		\centering
		\centerline{
			\begin{subfigure}[b]{.46\textwidth}
				{\resizebox{0.44\columnwidth}{!}{
%
%
\begin{tikzpicture}

\begin{axis}[%
width=4.047in,
height=3.566in,
at={(0.679in,0.481in)},
scale only axis,
point meta min=32.25,
point meta max=49.51,
axis on top,
xmin=0.5,
xmax=3.5,
xtick={1,2,3},
xticklabels={{N.},{R.},{O.}},
xlabel style={font=\color{white!15!black}},
xlabel={\Huge $\rightarrow\,$Target},
ylabel={\Huge Source$\,\rightarrow$},
y dir=reverse,
ymin=0.5,
ymax=4.5,
ytick={1,2,3,4},
yticklabels={{N.},{R.},{O.},{M.}},
ylabel style={font=\color{white!15!black}},
yticklabel style = {font=\Huge},
xticklabel style = {font=\Huge, yshift=-1mm},
title = {\Huge RTNH},
axis background/.style={fill=white},
colormap={mymap}{[1pt] rgb(0pt)=(0,0,0); rgb(204pt)=(1,0.62496,0.398); rgb(255pt)=(1,0.7812,0.4975)},
]
\addplot [forget plot] graphics [xmin=0.5, xmax=3.5, ymin=0.5, ymax=4.5] {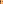};
\node[centered, align=center, inner sep=0, font=\color{white}]
at (axis cs:1,1) {\huge \textbf{40.8}};
\node[centered, align=center, inner sep=0, font=\color{white}]
at (axis cs:2,1) {\huge \textbf{49.0}};
\node[centered, align=center, inner sep=0, font=\color{white}]
at (axis cs:3,1) {\huge \textbf{48.9}};
\node[centered, align=center, inner sep=0, font=\color{white}]
at (axis cs:1,2) {\huge \textbf{36.5}};
\node[centered, align=center, inner sep=0, font=\color{white}]
at (axis cs:2,2) {\huge \textbf{48.8}};
\node[centered, align=center, inner sep=0, font=\color{white}]
at (axis cs:3,2) {\huge \textbf{41.8}};
\node[centered, align=center, inner sep=0, font=\color{white}]
at (axis cs:1,3) {\huge \textbf{40.4}};
\node[centered, align=center, inner sep=0, font=\color{white}]
at (axis cs:2,3) {\huge \textbf{49.5}};
\node[centered, align=center, inner sep=0, font=\color{white}]
at (axis cs:3,3) {\huge \textbf{48.8}};
\node[centered, align=center, inner sep=0, font=\color{white}]
at (axis cs:1,4) {\huge \textbf{40.2}};
\node[centered, align=center, inner sep=0, font=\color{white}]
at (axis cs:2,4) {\huge \textbf{48.6}};
\node[centered, align=center, inner sep=0, font=\color{white}]
at (axis cs:3,4) {\huge \textbf{46.3}};
\end{axis}
\end{tikzpicture}%
}}
				\hspace{1mm}
				{\resizebox{0.5145\columnwidth}{!}{
%
%
\begin{tikzpicture}

\begin{axis}[%
width=4.047in,
height=3.566in,
at={(0.679in,0.481in)},
scale only axis,
point meta min=32.25,
point meta max=49.51,
axis on top,
xmin=0.5,
xmax=3.5,
xtick={1,2,3},
xticklabels={{N.},{R.},{O.}},
xlabel style={font=\color{white!15!black}},
xlabel={\Huge $\rightarrow\,$Target},
ylabel={\Huge Source$\,\rightarrow$},
y dir=reverse,
ymin=0.5,
ymax=4.5,
ytick={1,2,3,4},
yticklabels={{N.},{R.},{O.},{M.}},
ylabel style={font=\color{white!15!black}},
yticklabel style = {font=\Huge},
xticklabel style = {font=\Huge,yshift=-1mm},
title = {\Huge SECOND},
axis background/.style={fill=white},
colormap={mymap}{[1pt] rgb(0pt)=(0,0,0); rgb(204pt)=(1,0.62496,0.398); rgb(255pt)=(1,0.7812,0.4975)},
colorbar,
colorbar style={font=\huge, ytick={32,36,40,44,48}, ytick style={draw=none},xshift=3mm}
]
\addplot [forget plot] graphics [xmin=0.5, xmax=3.5, ymin=0.5, ymax=4.5] {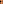};
\node[centered, align=center, inner sep=0, font=\color{white}]
at (axis cs:1,1) {\huge \textbf{40.5}};
\node[centered, align=center, inner sep=0, font=\color{white}]
at (axis cs:2,1) {\huge \textbf{45.3}};
\node[centered, align=center, inner sep=0, font=\color{white}]
at (axis cs:3,1) {\huge \textbf{42.5}};
\node[centered, align=center, inner sep=0, font=\color{white}]
at (axis cs:1,2) {\huge \textbf{32.2}};
\node[centered, align=center, inner sep=0, font=\color{white}]
at (axis cs:2,2) {\huge \textbf{41.5}};
\node[centered, align=center, inner sep=0, font=\color{white}]
at (axis cs:3,2) {\huge \textbf{40.6}};
\node[centered, align=center, inner sep=0, font=\color{white}]
at (axis cs:1,3) {\huge \textbf{39.8}};
\node[centered, align=center, inner sep=0, font=\color{white}]
at (axis cs:2,3) {\huge \textbf{45.5}};
\node[centered, align=center, inner sep=0, font=\color{white}]
at (axis cs:3,3) {\huge \textbf{42.2}};
\node[centered, align=center, inner sep=0, font=\color{white}]
at (axis cs:1,4) {\huge \textbf{39.5}};
\node[centered, align=center, inner sep=0, font=\color{white}]
at (axis cs:2,4) {\huge \textbf{44.5}};
\node[centered, align=center, inner sep=0, font=\color{white}]
at (axis cs:3,4) {\huge \textbf{42.1}};
\end{axis}
\end{tikzpicture}%
}}
				\caption{$AP_{BEV}$ Sedan}
				\label{fig:apbev_sedam}
			\end{subfigure}%
			\hfill
			\begin{subfigure}[b]{.46\textwidth}
				{\resizebox{0.44\columnwidth}{!}{
%
%
\begin{tikzpicture}

\begin{axis}[%
width=4.047in,
height=3.566in,
at={(0.679in,0.481in)},
scale only axis,
point meta min=13.58,
point meta max=28.75,
axis on top,
xmin=0.5,
xmax=3.5,
xtick={1,2,3},
xticklabels={{N.},{R.},{O.}},
xlabel style={font=\color{white!15!black}},
xlabel={\Huge $\rightarrow\,$Target},
ylabel={\Huge Source$\,\rightarrow$},
y dir=reverse,
ymin=0.5,
ymax=4.5,
ytick={1,2,3,4},
yticklabels={{N.},{R.},{O.},{M.}},
ylabel style={font=\color{white!15!black}},
yticklabel style = {font=\Huge},
xticklabel style = {font=\Huge,yshift=-1mm},
title = {\Huge RTNH},
axis background/.style={fill=white},
colormap={mymap}{[1pt] rgb(0pt)=(0,0,0); rgb(204pt)=(1,0.62496,0.398); rgb(255pt)=(1,0.7812,0.4975)},
]
\addplot [forget plot] graphics [xmin=0.5, xmax=3.5, ymin=0.5, ymax=4.5] {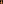};
\node[centered, align=center, inner sep=0, font=\color{white}]
at (axis cs:1,1) {\huge \textbf{16.5}};
\node[centered, align=center, inner sep=0, font=\color{white}]
at (axis cs:2,1) {\huge \textbf{20.1}};
\node[centered, align=center, inner sep=0, font=\color{white}]
at (axis cs:3,1) {\huge \textbf{16.3}};
\node[centered, align=center, inner sep=0, font=\color{white}]
at (axis cs:1,2) {\huge \textbf{13.6}};
\node[centered, align=center, inner sep=0, font=\color{white}]
at (axis cs:2,2) {\huge \textbf{22.7}};
\node[centered, align=center, inner sep=0, font=\color{white}]
at (axis cs:3,2) {\huge \textbf{14.7}};
\node[centered, align=center, inner sep=0, font=\color{white}]
at (axis cs:1,3) {\huge \textbf{19.2}};
\node[centered, align=center, inner sep=0, font=\color{white}]
at (axis cs:2,3) {\huge \textbf{25.0}};
\node[centered, align=center, inner sep=0, font=\color{white}]
at (axis cs:3,3) {\huge \textbf{16.6}};
\node[centered, align=center, inner sep=0, font=\color{white}]
at (axis cs:1,4) {\huge \textbf{16.8}};
\node[centered, align=center, inner sep=0, font=\color{white}]
at (axis cs:2,4) {\huge \textbf{22.9}};
\node[centered, align=center, inner sep=0, font=\color{white}]
at (axis cs:3,4) {\huge \textbf{17.8}};
\end{axis}
\end{tikzpicture}%
}}
				\hspace{1mm}
				{\resizebox{0.5145\columnwidth}{!}{
%
%
\begin{tikzpicture}

\begin{axis}[%
width=4.047in,
height=3.566in,
at={(0.679in,0.481in)},
scale only axis,
point meta min=13.58,
point meta max=28.75,
axis on top,
xmin=0.5,
xmax=3.5,
xtick={1,2,3},
xticklabels={{N.},{R.},{O.}},
xlabel style={font=\color{white!15!black}},
xlabel={\Huge $\rightarrow\,$Target},
ylabel={\Huge Source$\,\rightarrow$},
y dir=reverse,
ymin=0.5,
ymax=4.5,
ytick={1,2,3,4},
yticklabels={{N.},{R.},{O.},{M.}},
ylabel style={font=\color{white!15!black}},
yticklabel style = {font=\Huge},
xticklabel style = {font=\Huge,yshift=-1mm},
title = {\Huge SECOND},
axis background/.style={fill=white},
colormap={mymap}{[1pt] rgb(0pt)=(0,0,0); rgb(204pt)=(1,0.62496,0.398); rgb(255pt)=(1,0.7812,0.4975)},
colorbar,
colorbar style={font=\huge, ytick={16,20,24,28}, ytick style={draw=none},xshift=3mm}
]
\addplot [forget plot] graphics [xmin=0.5, xmax=3.5, ymin=0.5, ymax=4.5] {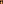};
\node[centered, align=center, inner sep=0, font=\color{white}]
at (axis cs:1,1) {\huge \textbf{20.6}};
\node[centered, align=center, inner sep=0, font=\color{white}]
at (axis cs:2,1) {\huge \textbf{21.7}};
\node[centered, align=center, inner sep=0, font=\color{white}]
at (axis cs:3,1) {\huge \textbf{19.1}};
\node[centered, align=center, inner sep=0, font=\color{white}]
at (axis cs:1,2) {\huge \textbf{14.0}};
\node[centered, align=center, inner sep=0, font=\color{white}]
at (axis cs:2,2) {\huge \textbf{20.9}};
\node[centered, align=center, inner sep=0, font=\color{white}]
at (axis cs:3,2) {\huge \textbf{16.5}};
\node[centered, align=center, inner sep=0, font=\color{white}]
at (axis cs:1,3) {\huge \textbf{19.5}};
\node[centered, align=center, inner sep=0, font=\color{white}]
at (axis cs:2,3) {\huge \textbf{28.8}};
\node[centered, align=center, inner sep=0, font=\color{white}]
at (axis cs:3,3) {\huge \textbf{18.4}};
\node[centered, align=center, inner sep=0, font=\color{white}]
at (axis cs:1,4) {\huge \textbf{17.8}};
\node[centered, align=center, inner sep=0, font=\color{white}]
at (axis cs:2,4) {\huge \textbf{25.9}};
\node[centered, align=center, inner sep=0, font=\color{white}]
at (axis cs:3,4) {\huge \textbf{19.8}};
\end{axis}
\end{tikzpicture}%
}}
				\caption{$AP_{BEV}$ Bus}
				\label{fig:apbev_bus}
			\end{subfigure}
		}
		\vspace{1.0mm}
		\centerline{
			\begin{subfigure}[b]{.46\textwidth}
				{\resizebox{0.44\columnwidth}{!}{
%
%
\begin{tikzpicture}

\begin{axis}[%
width=4.047in,
height=3.566in,
at={(0.679in,0.481in)},
scale only axis,
point meta min=32.25,
point meta max=49.51,
axis on top,
xmin=0.5,
xmax=3.5,
xtick={1,2,3},
xticklabels={{N.},{R.},{O.}},
xlabel style={font=\color{white!15!black}},
xlabel={\Huge $\rightarrow\,$Target},
ylabel={\Huge Source$\,\rightarrow$},
y dir=reverse,
ymin=0.5,
ymax=4.5,
ytick={1,2,3,4},
yticklabels={{N.},{R.},{O.},{M.}},
ylabel style={font=\color{white!15!black}},
yticklabel style = {font=\Huge},
xticklabel style = {font=\Huge, yshift=-1mm},
title = {\Huge RTNH},
axis background/.style={fill=white},
colormap={mymap}{[1pt] rgb(0pt)=(0,0,0); rgb(204pt)=(1,0.62496,0.398); rgb(255pt)=(1,0.7812,0.4975)},
]
\addplot [forget plot] graphics [xmin=0.5, xmax=3.5, ymin=0.5, ymax=4.5] {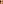};
\node[centered, align=center, inner sep=0, font=\color{white}]
at (axis cs:1,1) {\huge \textbf{39.2}};
\node[centered, align=center, inner sep=0, font=\color{white}]
at (axis cs:2,1) {\huge \textbf{47.7}};
\node[centered, align=center, inner sep=0, font=\color{white}]
at (axis cs:3,1) {\huge \textbf{42.0}};
\node[centered, align=center, inner sep=0, font=\color{white}]
at (axis cs:1,2) {\huge \textbf{31.8}};
\node[centered, align=center, inner sep=0, font=\color{white}]
at (axis cs:2,2) {\huge \textbf{41.8}};
\node[centered, align=center, inner sep=0, font=\color{white}]
at (axis cs:3,2) {\huge \textbf{40.5}};
\node[centered, align=center, inner sep=0, font=\color{white}]
at (axis cs:1,3) {\huge \textbf{38.1}};
\node[centered, align=center, inner sep=0, font=\color{white}]
at (axis cs:2,3) {\huge \textbf{47.9}};
\node[centered, align=center, inner sep=0, font=\color{white}]
at (axis cs:3,3) {\huge \textbf{41.8}};
\node[centered, align=center, inner sep=0, font=\color{white}]
at (axis cs:1,4) {\huge \textbf{38.4}};
\node[centered, align=center, inner sep=0, font=\color{white}]
at (axis cs:2,4) {\huge \textbf{43.2}};
\node[centered, align=center, inner sep=0, font=\color{white}]
at (axis cs:3,4) {\huge \textbf{42.0}};
\end{axis}
\end{tikzpicture}%
}}
				\hspace{1mm}
				{\resizebox{0.5145\columnwidth}{!}{
%
%
\begin{tikzpicture}

\begin{axis}[%
width=4.047in,
height=3.566in,
at={(0.679in,0.481in)},
scale only axis,
point meta min=32.25,
point meta max=49.51,
axis on top,
xmin=0.5,
xmax=3.5,
xtick={1,2,3},
xticklabels={{N.},{R.},{O.}},
xlabel style={font=\color{white!15!black}},
xlabel={\Huge $\rightarrow\,$Target},
ylabel={\Huge Source$\,\rightarrow$},
y dir=reverse,
ymin=0.5,
ymax=4.5,
ytick={1,2,3,4},
yticklabels={{N.},{R.},{O.},{M.}},
ylabel style={font=\color{white!15!black}},
yticklabel style = {font=\Huge},
xticklabel style = {font=\Huge,yshift=-1mm},
title = {\Huge SECOND},
axis background/.style={fill=white},
colormap={mymap}{[1pt] rgb(0pt)=(0,0,0); rgb(204pt)=(1,0.62496,0.398); rgb(255pt)=(1,0.7812,0.4975)},
colorbar,
colorbar style={font=\huge, ytick={32,36,40,44,48}, ytick style={draw=none},xshift=3mm}
]
\addplot [forget plot] graphics [xmin=0.5, xmax=3.5, ymin=0.5, ymax=4.5] {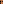};
\node[centered, align=center, inner sep=0, font=\color{white}]
at (axis cs:1,1) {\huge \textbf{38.9}};
\node[centered, align=center, inner sep=0, font=\color{white}]
at (axis cs:2,1) {\huge \textbf{41.2}};
\node[centered, align=center, inner sep=0, font=\color{white}]
at (axis cs:3,1) {\huge \textbf{41.4}};
\node[centered, align=center, inner sep=0, font=\color{white}]
at (axis cs:1,2) {\huge \textbf{30.7}};
\node[centered, align=center, inner sep=0, font=\color{white}]
at (axis cs:2,2) {\huge \textbf{40.2}};
\node[centered, align=center, inner sep=0, font=\color{white}]
at (axis cs:3,2) {\huge \textbf{33.8}};
\node[centered, align=center, inner sep=0, font=\color{white}]
at (axis cs:1,3) {\huge \textbf{35.5}};
\node[centered, align=center, inner sep=0, font=\color{white}]
at (axis cs:2,3) {\huge \textbf{41.3}};
\node[centered, align=center, inner sep=0, font=\color{white}]
at (axis cs:3,3) {\huge \textbf{40.9}};
\node[centered, align=center, inner sep=0, font=\color{white}]
at (axis cs:1,4) {\huge \textbf{32.7}};
\node[centered, align=center, inner sep=0, font=\color{white}]
at (axis cs:2,4) {\huge \textbf{43.0}};
\node[centered, align=center, inner sep=0, font=\color{white}]
at (axis cs:3,4) {\huge \textbf{40.9}};
\end{axis}
\end{tikzpicture}%
}}
				\caption{$AP_{3D}$ Sedan}
				\label{fig:ab3d_sedan}
			\end{subfigure}%
			\hfill
			\begin{subfigure}[b]{.46\textwidth}
				{\resizebox{0.44\columnwidth}{!}{
%
%
\begin{tikzpicture}

\begin{axis}[%
width=4.047in,
height=3.566in,
at={(0.679in,0.481in)},
scale only axis,
point meta min=13.58,
point meta max=28.75,
axis on top,
xmin=0.5,
xmax=3.5,
xtick={1,2,3},
xticklabels={{N.},{R.},{O.}},
xlabel style={font=\color{white!15!black}},
xlabel={\Huge $\rightarrow\,$Target},
ylabel={\Huge Source$\,\rightarrow$},
y dir=reverse,
ymin=0.5,
ymax=4.5,
ytick={1,2,3,4},
yticklabels={{N.},{R.},{O.},{M.}},
ylabel style={font=\color{white!15!black}},
yticklabel style = {font=\Huge},
xticklabel style = {font=\Huge,yshift=-1mm},
title = {\Huge RTNH},
axis background/.style={fill=white},
colormap={mymap}{[1pt] rgb(0pt)=(0,0,0); rgb(204pt)=(1,0.62496,0.398); rgb(255pt)=(1,0.7812,0.4975)},
]
\addplot [forget plot] graphics [xmin=0.5, xmax=3.5, ymin=0.5, ymax=4.5] {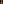};
\node[centered, align=center, inner sep=0, font=\color{white}]
at (axis cs:1,1) {\huge \textbf{15.0}};
\node[centered, align=center, inner sep=0, font=\color{white}]
at (axis cs:2,1) {\huge \textbf{19.2}};
\node[centered, align=center, inner sep=0, font=\color{white}]
at (axis cs:3,1) {\huge \textbf{15.6}};
\node[centered, align=center, inner sep=0, font=\color{white}]
at (axis cs:1,2) {\huge \textbf{12.9}};
\node[centered, align=center, inner sep=0, font=\color{white}]
at (axis cs:2,2) {\huge \textbf{19.5}};
\node[centered, align=center, inner sep=0, font=\color{white}]
at (axis cs:3,2) {\huge \textbf{13.6}};
\node[centered, align=center, inner sep=0, font=\color{white}]
at (axis cs:1,3) {\huge \textbf{16.1}};
\node[centered, align=center, inner sep=0, font=\color{white}]
at (axis cs:2,3) {\huge \textbf{22.0}};
\node[centered, align=center, inner sep=0, font=\color{white}]
at (axis cs:3,3) {\huge \textbf{15.7}};
\node[centered, align=center, inner sep=0, font=\color{white}]
at (axis cs:1,4) {\huge \textbf{15.8}};
\node[centered, align=center, inner sep=0, font=\color{white}]
at (axis cs:2,4) {\huge \textbf{19.4}};
\node[centered, align=center, inner sep=0, font=\color{white}]
at (axis cs:3,4) {\huge \textbf{16.8}};
\end{axis}
\end{tikzpicture}%
}}
				\hspace{1mm}
				{\resizebox{0.5145\columnwidth}{!}{
%
%
\begin{tikzpicture}

\begin{axis}[%
width=4.047in,
height=3.566in,
at={(0.679in,0.481in)},
scale only axis,
point meta min=13.58,
point meta max=28.75,
axis on top,
xmin=0.5,
xmax=3.5,
xtick={1,2,3},
xticklabels={{N.},{R.},{O.}},
xlabel style={font=\color{white!15!black}},
xlabel={\Huge $\rightarrow\,$Target},
ylabel={\Huge Source$\,\rightarrow$},
y dir=reverse,
ymin=0.5,
ymax=4.5,
ytick={1,2,3,4},
yticklabels={{N.},{R.},{O.},{M.}},
ylabel style={font=\color{white!15!black}},
yticklabel style = {font=\Huge},
xticklabel style = {font=\Huge,yshift=-1mm},
title = {\Huge SECOND},
axis background/.style={fill=white},
colormap={mymap}{[1pt] rgb(0pt)=(0,0,0); rgb(204pt)=(1,0.62496,0.398); rgb(255pt)=(1,0.7812,0.4975)},
colorbar,
colorbar style={font=\huge, ytick={16,20,24,28}, ytick style={draw=none},xshift=3mm}
]
\addplot [forget plot] graphics [xmin=0.5, xmax=3.5, ymin=0.5, ymax=4.5] {fig/tikz_heatmaps/AP_BEV_Bus_Second-1.png};
\node[centered, align=center, inner sep=0, font=\color{white}]
at (axis cs:1,1) {\huge \textbf{18.5}};
\node[centered, align=center, inner sep=0, font=\color{white}]
at (axis cs:2,1) {\huge \textbf{21.0}};
\node[centered, align=center, inner sep=0, font=\color{white}]
at (axis cs:3,1) {\huge \textbf{17.9}};
\node[centered, align=center, inner sep=0, font=\color{white}]
at (axis cs:1,2) {\huge \textbf{13.0}};
\node[centered, align=center, inner sep=0, font=\color{white}]
at (axis cs:2,2) {\huge \textbf{20.1}};
\node[centered, align=center, inner sep=0, font=\color{white}]
at (axis cs:3,2) {\huge \textbf{15.7}};
\node[centered, align=center, inner sep=0, font=\color{white}]
at (axis cs:1,3) {\huge \textbf{18.6}};
\node[centered, align=center, inner sep=0, font=\color{white}]
at (axis cs:2,3) {\huge \textbf{27.9}};
\node[centered, align=center, inner sep=0, font=\color{white}]
at (axis cs:3,3) {\huge \textbf{17.4}};
\node[centered, align=center, inner sep=0, font=\color{white}]
at (axis cs:1,4) {\huge \textbf{16.8}};
\node[centered, align=center, inner sep=0, font=\color{white}]
at (axis cs:2,4) {\huge \textbf{22.7}};
\node[centered, align=center, inner sep=0, font=\color{white}]
at (axis cs:3,4) {\huge \textbf{18.8}};
\end{axis}
\end{tikzpicture}%
}}
				\caption{$AP_{3D}$ Bus}
				\label{fig:ab3d_bus}
			\end{subfigure}
		}
		\caption{Evaluation of $AP_{BEV}$ and $AP_{3D}$ performance metrics for weather domain shift on Bosch-Radar data with 60k training samples (source$\rightarrow$target). \textbf{N.}:Normal, \textbf{R.}:Rain, \textbf{O.}:Overcast, \textbf{M.}:Mixed.}
		\label{fig: heatmap}
		\vspace{-7mm}
	\end{figure*}
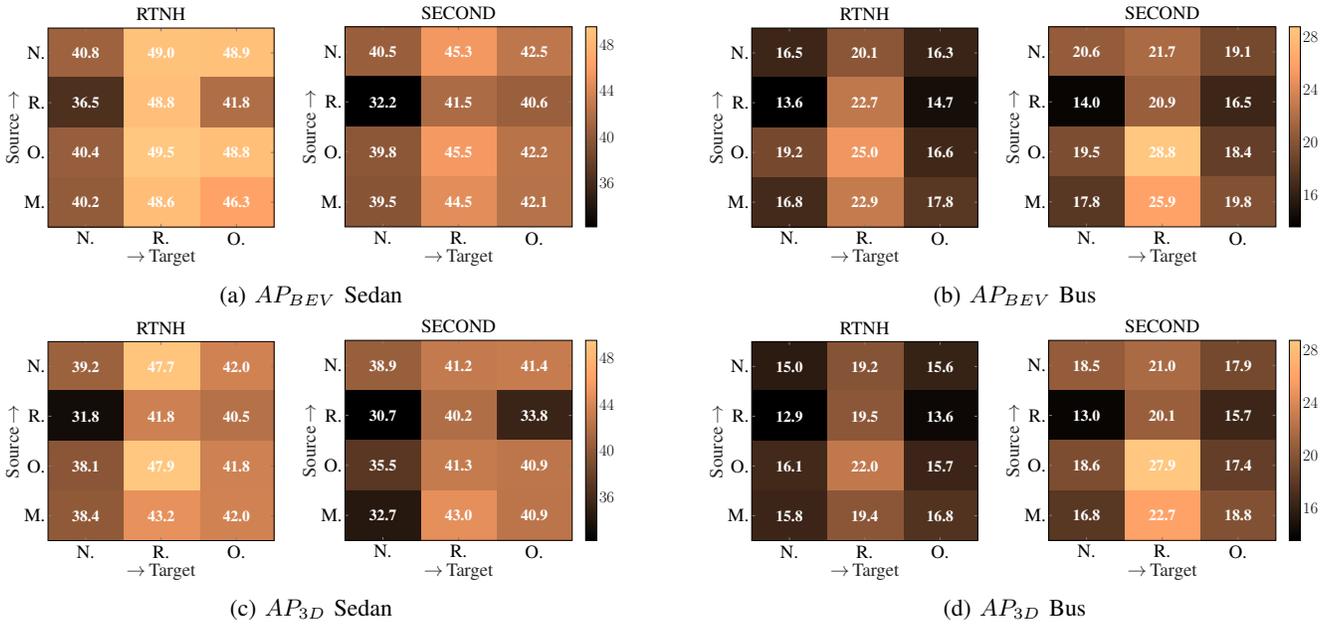

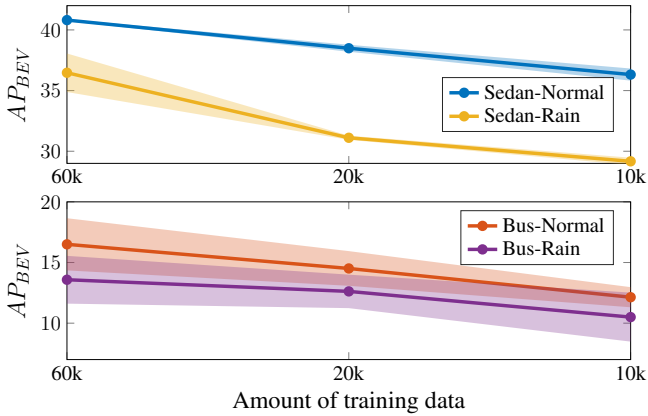
\begin{figure}[tbp]
		\vspace{1em}
		\centering
		{\resizebox{\columnwidth}{!}{
%
%
\definecolor{mycolor1}{rgb}{0.00000,0.44700,0.74100}%
\definecolor{mycolor2}{rgb}{0.85000,0.32500,0.09800}%
\definecolor{mycolor3}{rgb}{0.92900,0.69400,0.12500}%
\definecolor{mycolor4}{rgb}{0.49400,0.18400,0.55600}%
\begin{tikzpicture}

\begin{axis}[%
width=4.521in,
height=1.266in,
at={(0.758in,0.481in)},
scale only axis,
xmin=1,
xmax=3,
xtick={1,2,3},
xticklabels={{60k},{20k},{10k}},
xlabel style={font=\color{white!15!black}},
ymin=29,
ymax=42,
ylabel style={font=\color{white!15!black}},
ylabel={\Large $AP_{BEV}$},
yticklabel style = {font=\large},
xticklabel style = {font=\large},
axis background/.style={fill=white},
axis x line*=bottom,
axis y line*=left,
axis lines = box,
legend style={at={(0.97,0.2)}, anchor=south east, legend cell align=left, align=left, draw=white!15!black}
]

\addplot[area legend, draw=black, fill=mycolor1, draw opacity=0, fill opacity=0.3, forget plot]
table[row sep=crcr] {%
x	y\\
1	40.73\\
2	38.2\\
3	35.82\\
3	36.82\\
2	38.78\\
1	40.89\\
}--cycle;
\addplot [color=mycolor1, line width=2.0pt, mark=*, mark options={solid, fill=mycolor1, mycolor1}]
  table[row sep=crcr]{%
1	40.81\\
2	38.49\\
3	36.32\\
};
\addlegendentry{\large Sedan-Normal}

\addplot[area legend, draw=black, fill=mycolor3, draw opacity=0, fill opacity=0.3, forget plot]
table[row sep=crcr] {%
x	y\\
1	34.88\\
2	30.94\\
3	28.87\\
3	29.47\\
2	31.28\\
1	38.06\\
}--cycle;
\addplot [color=mycolor3, line width=2.0pt, mark=*, mark options={solid, fill=mycolor3, mycolor3}]
  table[row sep=crcr]{%
1	36.47\\
2	31.11\\
3	29.17\\
};
\addlegendentry{\large Sedan-Rain}

\end{axis}
\end{tikzpicture}%
}}
		\vspace{1mm}
		{\resizebox{\columnwidth}{!}{
%
%
\definecolor{mycolor1}{rgb}{0.00000,0.44700,0.74100}%
\definecolor{mycolor2}{rgb}{0.85000,0.32500,0.09800}%
\definecolor{mycolor3}{rgb}{0.92900,0.69400,0.12500}%
\definecolor{mycolor4}{rgb}{0.49400,0.18400,0.55600}%
\begin{tikzpicture}

\begin{axis}[%
width=4.521in,
height=1.266in,
at={(0.758in,0.481in)},
scale only axis,
xmin=1,
xmax=3,
xtick={1,2,3},
xticklabels={{60k},{20k},{10k}},
xlabel style={font=\color{white!15!black}},
xlabel={\Large Amount of training data},
ymin=7,
ymax=20,
ylabel style={font=\color{white!15!black}},
ylabel={\Large $AP_{BEV}$},
yticklabel style = {font=\large},
xticklabel style = {font=\large},
axis background/.style={fill=white},
axis x line*=bottom,
axis y line*=left,
axis lines = box,
legend style={at={(0.97,0.6)}, anchor=south east, legend cell align=left, align=left, draw=white!15!black}
]

\addplot[area legend, draw=black, fill=mycolor2, draw opacity=0, fill opacity=0.3, forget plot]
table[row sep=crcr] {%
	x	y\\
	1	14.35\\
	2	13.08\\
	3	11.31\\
	3	12.97\\
	2	15.94\\
	1	18.65\\
}--cycle;
\addplot [color=mycolor2, line width=2.0pt, mark=*, mark options={solid, fill=mycolor2, mycolor2}]
table[row sep=crcr]{%
	1	16.5\\
	2	14.51\\
	3	12.14\\
};
\addlegendentry{\large Bus-Normal}

\addplot[area legend, draw=black, fill=mycolor4, draw opacity=0, fill opacity=0.3, forget plot]
table[row sep=crcr] {%
	x	y\\
	1	11.61\\
	2	11.24\\
	3	8.48\\
	3	12.52\\
	2	14\\
	1	15.55\\
}--cycle;
\addplot [color=mycolor4, line width=2.0pt, mark=*, mark options={solid, fill=mycolor4, mycolor4}]
table[row sep=crcr]{%
	1	13.58\\
	2	12.62\\
	3	10.5\\
};
\addlegendentry{\large Bus-Rain}

\end{axis}
\end{tikzpicture}%
}}
		\caption{$AP_{BEV}$ under normal and rainy conditions across varying training data sizes.}
		\label{fig: scale}
		\vspace{-6mm}
	\end{figure}

	As introduced in Sec.~\ref{sec: dataset}, the Bosch-Radar dataset is considerably larger than the K-Radar dataset. We utilize this dataset to investigate the domain shift effect across three weather conditions: normal, overcast, and rain. Following the protocols established with K-Radar, we conduct a two-class object detection task using the same two networks. For clarity, the results on 60k scales are visualized in Fig.~\ref{fig: heatmap}. The key findings are listed below:
	\begin{enumerate}[label=\arabic*., wide=0.1em, itemsep=0.25em]
		\item \textit{When examining the rain performance, we notice a significant domain shift effect from rain to normal and overcast conditions}: For example, on RTNH model, there is a decrease of $7.4\%$ in $AP_{3D}$ under normal conditions and a $7.1\%$ decrease in $AP_{BEV}$ under overcast conditions for class sedan comparing the rain-trained model with the source domain-trained models. Conversely, models trained under normal or overcast conditions still adapt well to rain patterns: the results are slightly better than the rain-trained model in terms of $AP_{BEV}$ and better, by around $6\%$, in $AP_{3D}$. This aligns with observations from lidar domain research~\cite{eskandar2024empirical}, which shows that normal weather-trained data can already give competitively strong results in adverse weather conditions.  
		
		\item \textit{The model trained on a mixture of datasets does not exhibit any superiority compared to models trained with other data sets, despite the increased diversity of its training inputs.} We can tell from the figure that all the results from mixed-trained models have competitive or worse results than the normal/overcast-trained models. This observation is consistent with findings from lidar-related research~\cite{vattem2022rethinking}. 
		Based on the observation here, we claim that the superiority of multi-domain over single-domain~\cite{wu2023towards} is not applicable to the weather domain shift problem.  
		
		\item \textit{Solely increasing the dataset size may not effectively mitigate the effects of domain shift.} This observation is based on comparisons among models trained with 60k, 20k, and 10k samples. The impact of weather domain shifts appears consistent (as shown in Fig.~\ref{fig: scale}). Models trained on data from rainy conditions consistently performed worse than other models. Additionally, the domain gap remained uniform for both classes. This observation suggests that solely increasing the size of the dataset may not effectively mitigate the effects of domain shift from rain to normal. Though acquiring normal weather data is easier than rain data, investigating this issue for model behavior insights is still valuable.
	\end{enumerate}
	
	\subsection{Comparison}
	By comparing the domain gaps in the two datasets, \textit{we find that weather domain shift effects vary between different datasets.} As discussed in the previous two subsections, the impacts of domain shifts on the K-Radar and Bosch-Radar datasets differ significantly. Except for overcast conditions, all other adverse weather scenarios in the K-Radar data experience significant domain shift effects. In contrast, the Bosch-Radar data only exhibits adverse effects when transitioning from adverse to favorable weather conditions. Remarkably, models trained on favorable weather data often outperform those trained on adverse weather data.
	
	Potential reasons for these variations are as follows: First, the K-Radar dataset may not be sufficiently representative due to its mid-size. Second, and crucially, unlike lidar, which directly obtains point clouds from the sensor, the point clouds from the radar are derived through signal processing on spectrum data. The point cloud can significantly differ due to the sensor type and processing techniques. Here, the two datasets are using different methods to generate their point clouds: The method of K-Radar only selects the top $k\%$ highest power point from the 4D radar cube and then converts them into the Cartesian coordinates as 3D points with the power value. Our Bosch-Radar data uses a peak detection method, such as CFAR~\cite{cfar}, to select relevant points. As a result, the point cloud distribution from the two datasets can be significantly different, which may cause a different domain shift effect. The disparity in our results indicates that the domain shift effect in radar data is highly specific to the datasets and is not universally applicable across all datasets.
	
	\section{Road Domain shift}
	In this section, we explore the impact of domain shift across various road types. Our experiment utilizes the Bosch-Radar dataset due to the large scale. Considering the varying numbers of objects in different road conditions mentioned in Sec.~\ref{sec: implement}, we use the same number of objects rather than frames across all subsets and focus exclusively on detecting a single class: ``sedan''. The results are shown in Table~\ref{tab:road}. We trained the model on four different road conditions: only urban, only rural, only highway, and a mix of all three types. The test set is fixed-chosen, containing all three different road types. Based on the result, we discover that:
	\begin{table}[bp]
		\vspace{-1em}
		\centering
		\caption{$AP_{BEV}$/$AP_{3D}$ comparison of road domain shift (source$\rightarrow$target) on Bosch-Radar Data with RTNH~\cite{paek2022k}. }
		\Huge
		\resizebox{\columnwidth}{!}{
			\label{tab:road}
			\begin{threeparttable}
				\renewcommand{\arraystretch}{1.5}
				\begin{tabular}{l|ccc}
					\toprule
					Sedan& $\rightarrow$Urban & $\rightarrow$Rural &  $\rightarrow$Highway\\
					\hline
					Urban$\rightarrow$ & $\textbf{40.16} \pm \scriptstyle{0.09}/\textbf{38.64} \pm \scriptstyle{0.13}$ & $\textbf{54.78} \pm \scriptstyle{0.15}/\textcolor{red}{\textbf{47.75} \pm \scriptstyle{0.15}}$ &  $\textbf{47.45} \pm \scriptstyle{0.20}/\textbf{45.76} \pm \scriptstyle{0.16}$ \\
					Rural$\rightarrow$ & $\textbf{39.33} \pm \scriptstyle{0.11}/\textbf{37.10} \pm \scriptstyle{0.15}$ &  $\textbf{56.67} \pm \scriptstyle{0.06}/\textbf{53.57} \pm \scriptstyle{0.14}$& $\textbf{49.35} \pm \scriptstyle{0.06}/\textbf{48.07} \pm \scriptstyle{0.12}$\\
					Highway$\rightarrow$ & \textcolor{red}{$\textbf{31.51} \pm \scriptstyle{0.02}/\textbf{30.31} \pm \scriptstyle{0.01}$}& $\textbf{53.85} \pm \scriptstyle{2.77}/\textbf{\textcolor{red}{48.49}} \pm \scriptstyle{0.08}$ & $\textbf{49.58} \pm \scriptstyle{0.16}/\textbf{48.35} \pm \scriptstyle{0.22}$\\
					Mixed$\rightarrow$&$\textbf{39.22} \pm \scriptstyle{0.06}/\textbf{35.25} \pm \scriptstyle{2.68}$ &$\textbf{56.11} \pm \scriptstyle{0.38}/\textbf{50.78} \pm \scriptstyle{2.74}$ & $\textbf{49.38} \pm \scriptstyle{0.01}/\textbf{47.84} \pm \scriptstyle{0.08}$\\
					\bottomrule
				\end{tabular}
				\begin{tablenotes}
					\item {Note: Noticeable performance drop ($5\%+$) is highlighted in \textcolor{red}{RED} (\textbf{Comparison across rows}).}
				\end{tablenotes}
			\end{threeparttable}
		}
	\end{table}
	
	\begin{enumerate}[label=\arabic*., wide=0.1em, itemsep=0.25em]
		\item \textit{A unidirectional domain shift from highway to urban/rural occurs.} The performance of the model trained on highway data significantly decreased in urban areas ($-8.65\%~AP_{BEV}/-8.33\%~AP_{3D}$) and by approximately $5\%~AP_{3D}$ in rural areas. Conversely, the model trained on urban/rural data performs competitively in highway scenarios. Generally, the urban/rural environment is more complex than highways due to differences in traffic density and environmental context (such as buildings and roadside facilities).
		
		\item \textit{The urban-trained model underperformed on rural data, showing a drop of $-5.82\%~AP_{3D}$ compared to the rural-trained model.} 
		We hypothesize that the robustness of the rural-trained model across all conditions is due to its intermediate nature, incorporating both built-up areas like townships and sparse scenes like rural roads. Rural driving behavior is also more varied than in urban or highway settings. For instance, in Germany, the speed limit in urban areas is typically 50 km/h, while in rural areas, it can range from 50 km/h to 100 km/h. This diverse training data likely enhances the model's generalization ability and robustness, aligning with our observations.
		
		\item \textit{The mixed road data also can provide compatible results on each condition.} Our investigation revealed that the mixed-trained model did not experience a significant performance drop compared to the peak values, indicating its strong generalization ability. While domain shift occurs between different road conditions, it is not as pronounced as with weather conditions. This is because road conditions do not impact radar signal propagation as significantly as weather does. However, variations in environmental background and driving behavior can still influence domain shift. 
	\end{enumerate}
	Our experimental results demonstrate that collecting data from diverse road types is crucial for maintaining strong performance in complex scenarios, particularly in rural settings.
	
	\section{Limitations}
	While the study is detailed and comprehensive, it is not without limitations.
	First, as an empirical study, our conclusions are based on analyzing experimental results. However, a theoretical analysis of domain shift under varying environmental conditions still needs further research. Second, although our dataset is large-scale, it lacks data on highly adverse weather conditions, such as snow, which are also scarce in K-Radar. These conditions warrant further investigation. Additionally, our study specifically focuses on environmental domain shifts involving radar point cloud data and does not cover behavior in radar spectrum data. Other critical factors, such as sensor type, also play an essential role in practical applications and should be further explored.
	
	\section{Conclusion}
	This paper presents an empirical study on the domain shift effect in radar-based object detection under diverse environmental conditions. We investigated this phenomenon using two datasets with various configurations, including weather conditions, road conditions, and dataset scales. Our findings reveal that domain shifts do occur across different weather conditions, with the effect varying between datasets. Additionally, we observed domain shifts between different road types, particularly from highways to urban areas.
	
	To the best of our knowledge, our work is the first to investigate these effects within the context of 4D radar-based object detection. We hope our research will contribute to a deeper understanding of domain shifts in radar perception systems and provide valuable insights for optimizing data collection strategies. In future work, we will investigate domain adaptation techniques to address challenges posed by environmental variations.

	\addtolength{\textheight}{-12cm}   

\end{document}